\documentclass[runningheads]{llncs}

\usepackage[T1]{fontenc}
\usepackage{graphicx}
\usepackage{amsmath,amssymb,bbm}
\usepackage{booktabs}
\usepackage{url}
\usepackage{microtype}

\begin{document}


\title{PRISM: Differentiable Analysis-by-Synthesis for Fixel Recovery in Diffusion MRI}
\titlerunning{PRISM: Differentiable Fixel Recovery in dMRI}

\author{Mohamed Abouagour\inst{1}$^{(\ast)}$ \and Atharva Shah\inst{1} \and Eleftherios Garyfallidis\inst{1}}
\authorrunning{M. Abouagour et al.}
\institute{Indiana University, Bloomington, IN, USA\\
$^{(\ast)}$Corresponding author: \email{moabouag@iu.edu}}

\maketitle

\begin{abstract}
Diffusion MRI microstructure fitting is nonconvex and often performed voxelwise, which limits fiber peak recovery in narrow crossings.
This work introduces PRISM, a differentiable analysis-by-synthesis framework that fits an explicit multi-compartment forward model end-to-end over spatial patches.
The model combines cerebrospinal fluid (CSF), gray matter, up to $K$ white-matter fiber compartments (stick-and-zeppelin), and a restricted compartment, with explicit fiber directions and soft model selection via repulsion and sparsity priors.
PRISM supports a fast MSE objective and a Rician negative log-likelihood (NLL) that jointly learns~$\sigma$ without oracle information.
A lightweight nuisance calibration module (smooth bias field and per-measurement scale/offset) is included for robustness and regularized to identity in clean-data tests.
On synthetic crossing-fiber data (SNR\,=\,30; five methods, 16 crossing angles), PRISM achieves $3.5^\circ$ best-match angular error with 95\% recall, which is $1.9{\times}$ lower than the best baseline (MSMT-CSD, $6.8^\circ$, 83\% recall); in NLL mode with learned~$\sigma$, error drops to $2.3^\circ$ with 99\% recall, resolving crossings down to $20^\circ$.
On the DiSCo1 phantom (NLL mode), PRISM improves connectivity correlation over CSD baselines at all four tracking angles (best $r{=}.934$ at $25^\circ$ vs.\ $.920$ for MSMT-CSD).
Whole-brain HCP fitting (${\sim}$741k voxels, MSE mode) completes in ${\sim}$12\,min on a single GPU (hardware in Sec.~\ref{sec:experiments}) with near-identical results across random seeds.

\end{abstract}

\section{Introduction}

Diffusion MRI (dMRI) encodes tissue microstructure in direction- and $b$-value-dependent signal attenuation.
In practice, recovering multiple fiber populations from finite, noisy measurements is strongly nonconvex, and standard voxelwise fitting discards spatial coherence.
Intensity variations (bias field and per-measurement affine correction) and Rician magnitude noise further destabilize narrow-crossing recovery when handled in separate preprocessing stages.

\begin{figure}[t]
\centering
\includegraphics[width=\linewidth]{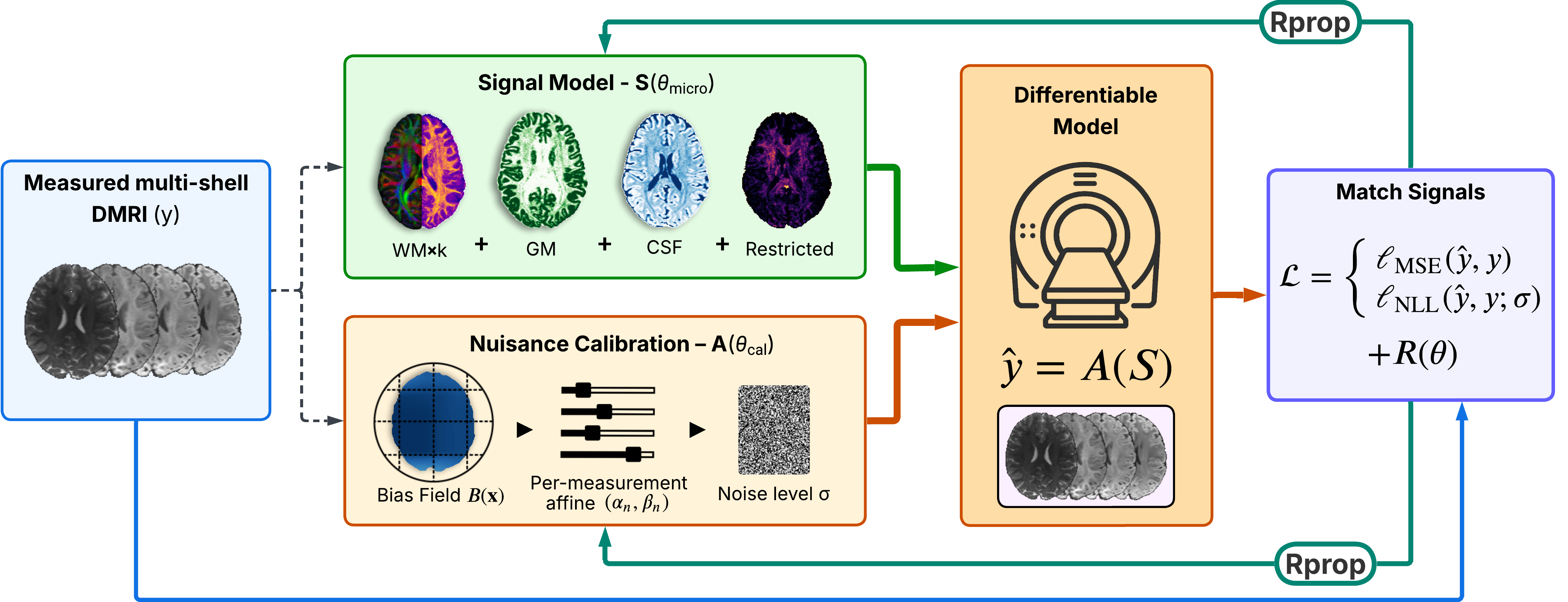}
\caption{\textbf{PRISM analysis-by-synthesis pipeline.}  The multi-compartment tissue signal model $\mathcal{S}(\boldsymbol{\theta}_{\mathrm{micro}})$ (top, green) combines up to $K$ white-matter fiber populations with gray matter, CSF, and a restricted compartment.  A nuisance calibration module $\mathcal{A}(\boldsymbol{\theta}_{\mathrm{cal}})$ (bottom, orange) applies a smooth bias field $B(\mathbf{x})$, per-measurement affine correction $(\alpha_n, \beta_n)$, and noise level~$\sigma$.  The differentiable forward model synthesizes the expected measurement $\hat{y}{=}\mathcal{A}(\mathcal{S})$, which is compared to observed multi-shell dMRI via an MSE or NLL loss plus regularization $\mathcal{R}(\boldsymbol{\theta})$.  Rprop updates tissue and calibration parameters jointly through gradient-based optimization.}
\label{fig:overview}
\end{figure}

This work proposes PRISM, a \emph{differentiable analysis-by-synthesis} framework for fitting microstructure and nuisance calibration parameters in one end-to-end optimization.
PRISM initializes tissue and calibration parameters, synthesizes the expected measurement through a differentiable forward model (Fig.~\ref{fig:overview}), and iteratively minimizes reconstruction error:
\begin{equation}
\hat{\mathbf{y}} = \underbrace{\mathcal{A}\!\bigl(\,\overbrace{\mathcal{S}(\boldsymbol{\theta}_{\mathrm{micro}})}^{\text{tissue signal}};\;\boldsymbol{\theta}_{\mathrm{cal}}\bigr)}_{\text{nuisance calibration}}, \qquad \min_{\boldsymbol{\theta}} \mathcal{L}(\hat{\mathbf{y}}, \mathbf{y}) + \mathcal{R}(\boldsymbol{\theta}),
\end{equation}
where $\mathcal{L}$ is either an MSE or Rician NLL data-fidelity term (Sec.~\ref{sec:optimization}).
Because every component is differentiable, gradients flow from reconstruction error through the calibration module to every tissue parameter, enabling the optimizer to focus on biological structure while absorbing residual nuisance.
Spatial priors further enforce coherence across voxels.

Multi-compartment models such as NODDI~\cite{zhang2012noddi}, Ball-and-Sticks~\cite{behrens2003characterization}, and multi-tissue CSD~\cite{jeurissen2014multi} typically fit voxelwise on preprocessed data.
PRISM instead combines interpretable biophysics with end-to-end optimization over volumes.

\noindent\textbf{Related work.}
Scalar microstructure models (DTI~\cite{basser1994dtimri}, DKI~\cite{jensen2005dki}) and orientation estimators (CSD~\cite{tournier2007csd}) assume artifact-free input and fit each voxel independently.
Deep-learning approaches learn voxelwise mappings from signals to microstructure~\cite{koppers2017spherical,nath2019deep,tian2020deepdti}, enabling fast inference but trading interpretability.
ReMiDi~\cite{khole2025remidi} builds a differentiable dMRI simulator for finite-element meshes and iterates a latent mesh representation to match observed signals.
Forward-simulation methods such as ODF-FP~\cite{baete2019odffp} and FORCE~\cite{shah2025force} precompute libraries of plausible fiber configurations and identify each voxel by nearest-neighbor lookup in signal or ODF space.
PRISM targets a different operating regime: the multi-compartment inverse problem on clinical-style acquisitions.  It uses an analytic signal model, optimizes over 3D spatial patches with neighbor-aware regularization, and jointly calibrates bias field, per-measurement affine correction, and noise level.

\noindent\textbf{Contributions.}
\textbf{(1)}~A differentiable multi-compartment microstructure fitter with explicit $K$-fiber parameterization, repulsion+sparsity-based soft model selection, and a restricted compartment.
\textbf{(2)}~Two data-fidelity modes: MSE for fast convergence, and Rician NLL with \emph{self-supervised} noise-level recovery, which lowers overall angular error from $3.5^\circ$ to $2.3^\circ$.
\textbf{(3)}~$1.9{\times}$ lower best-match angular error than the best baseline on a five-method crossing-fiber benchmark ($3.5^\circ$ vs.\ $6.8^\circ$ at SNR\,=\,30, MSE mode) with 95\% recall; NLL mode with learned~$\sigma$ reduces error to $2.3^\circ$ with 99\% recall; $+1.4$--$1.9$\,pp in tractography correlation on DiSCo1 (NLL).
\textbf{(4)}~Whole-brain fitting of $\sim$741k voxels in $\sim$12\,min on a single GPU with near-identical Dice/correlation across random seeds; optional nuisance calibration halves angular error under gain perturbations ($\sigma_g{=}0.20$) and self-regularizes to identity on unperturbed data.

\section{Methods}
\label{sec:methods}

\subsection{Analysis-by-Synthesis Framework}
PRISM fits a differentiable forward model to multi-shell dMRI data $\mathbf{y}$ by jointly optimizing tissue parameters $\boldsymbol{\theta}_{\mathrm{micro}}$ and nuisance calibration parameters $\boldsymbol{\theta}_{\mathrm{cal}}$ until the synthesized measurement matches the observed data.
At voxel $\mathbf{x}$ and measurement $n$, the forward model composes a tissue signal with a lightweight calibration cascade:
\begin{equation}
\hat{y}(\mathbf{x}, n) = \mathcal{A}\!\bigl(\mathcal{S}(\mathbf{x}, n;\, \boldsymbol{\theta}_{\mathrm{micro}});\, \boldsymbol{\theta}_{\mathrm{cal}}\bigr)
\label{eq:forward}
\end{equation}
where $\mathcal{S}$ generates the clean tissue signal and $\mathcal{A}$ applies nuisance intensity/noise calibration (Sec.~\ref{sec:artifact_model}).
All components are differentiable, so standard gradient-based optimization can update tissue and calibration parameters jointly.

\subsection{Tissue Signal Model}
\label{sec:tissue_model}
The multi-compartment signal combines cerebrospinal fluid (CSF), gray matter (GM), white matter (WM, up to $K$ fibers), and a restricted compartment, with fractions constrained via softmax:
\begin{equation}
\mathcal{S}(\mathbf{x}, n) = S_0(\mathbf{x})\!\left(f_{\mathrm{csf}}\,E_{\mathrm{csf}} + f_{\mathrm{gm}}\,E_{\mathrm{gm}} + \sum_{k=1}^{K} f_{\mathrm{wm},k}\,E_{\mathrm{wm},k} + f_{\mathrm{res}}\,E_{\mathrm{res}}\right)
\end{equation}
Isotropic compartments follow a mono-exponential decay with fixed diffusivities ($\times 10^{-3}$\,mm$^2$/s):
\begin{equation}
E_{\mathrm{csf}} = e^{-bD_{\mathrm{csf}}},\quad
E_{\mathrm{gm}} = e^{-bD_{\mathrm{gm}}},\quad
E_{\mathrm{res}} = e^{-bD_{\mathrm{res}}},
\end{equation}
where $D_{\mathrm{csf}}{=}3.0$, $D_{\mathrm{gm}}{=}0.9$, and $D_{\mathrm{res}}{=}0.2$.
Each WM fiber uses a stick-and-zeppelin model~\cite{behrens2003characterization}:
\begin{equation}
E_{\mathrm{wm}} = f_{\mathrm{intra}}\, e^{-bD_\parallel\cos^2\!\theta} + (1{-}f_{\mathrm{intra}})\, e^{-b(D_\parallel\cos^2\!\theta + D_\perp\sin^2\!\theta)}
\end{equation}
where $\theta$ is the angle between the fiber direction $\mathbf{d}_k$ and gradient direction $\mathbf{g}_n$, with fixed axial and radial diffusivities $D_\parallel{=}1.7$, $D_\perp{=}0.4$ ($\times 10^{-3}$\,mm$^2$/s), and learnable intra-axonal fraction $f_\mathrm{intra}$ (initialized at 0.5 via sigmoid).


\textbf{Restricted compartment.}
A restricted isotropic compartment ($D_\mathrm{res}{=}0.2$) captures slowly decaying signal at high $b$-values from water in small cellular structures (e.g., soma, astrocytes)~\cite{palombo2020sandi} unexplained by extra-cellular or free-water diffusion.

\textbf{Per-voxel parameters:} $S_0$ (softplus), $(K{+}3)$ tissue fractions (softmax over CSF, GM, $K$ WM fibers, and restricted), $K{\times}3$ fiber directions ($\ell_2$-normalized), and $f_\mathrm{intra}$ (sigmoid).

\subsection{Nuisance Intensity and Noise Calibration}
\label{sec:artifact_model}
The calibration module $\mathcal{A}$ applies three operators sequentially to the clean tissue signal $\mathcal{S}$: a spatial bias field, per-measurement intensity correction, and a learned noise level.

\noindent\textbf{1.\ Spatial bias field} $B(\mathbf{x})$: models multiplicative intensity variation from coil sensitivity and $B_1$ inhomogeneity.
A low-resolution $8^3$ control grid $\mathbf{b}_\mathrm{lr}$ (initialized at zero) is trilinearly upsampled to the data resolution and exponentiated:\\
$B(\mathbf{x}) = \exp\bigl(\mathrm{upsample}(\mathbf{b}_\mathrm{lr})\bigr)$,
enforcing low-frequency structure by construction.

\noindent\textbf{2.\ Per-measurement intensity correction}:
A per-measurement affine model absorbs residual gradient-direction-dependent intensity variation:
$\hat{y}_n \mapsto \exp(\alpha_n) \cdot \hat{y}_n + \beta_n$,
where $\hat{y}_n$ is the predicted signal for measurement~$n$, and $\alpha_n$, $\beta_n$ are initialized at zero (identity transform) and regularized toward identity ($\alpha_n{\to}0$, $\beta_n{\to}0$) via $L_2$ penalty.
This captures multiplicative and additive per-volume drift that survives standard preprocessing.

\noindent\textbf{3.\ Noise level} $\sigma = \exp(\sigma_\mathrm{log})$~\cite{gudbjartsson1995rician}: a learnable parameter recovered jointly with all others, serving as the noise scale for the Rician-aware loss mode (Sec.~\ref{sec:optimization}).

\noindent Composing the three operators, the full calibration chain for measurement~$n$ at voxel~$\mathbf{x}$ is:
\begin{equation}
\hat{y}_n(\mathbf{x}) = \exp(\alpha_n)\,B(\mathbf{x})\,\mathcal{S}_n(\mathbf{x}) + \beta_n.
\label{eq:calibration_chain}
\end{equation}

\noindent\textbf{Scale control.}
The product $B(\mathbf{x}) \cdot S_0(\mathbf{x})$ is scale-ambiguous; the $8^3$ control grid restricts $B$ to low spatial frequencies while $S_0$ captures per-voxel intensity, giving a well-posed decomposition.
$L_2$ penalties on $\mathbf{b}_\mathrm{lr}$ keep $B$ near unity, and the per-measurement affine ($\alpha_n$, $\beta_n$) is similarly regularized toward identity.

\subsection{Regularization}
\label{sec:regularization}
The joint tissue--calibration model is inherently underdetermined: multiple configurations can explain the same observations.
PRISM resolves this with lightweight regularizers that steer the solution toward anatomically plausible configurations without sacrificing data-driven flexibility.
Regularizer weights are fixed globally when enabled; benchmark-specific subsets are noted in Sec.~\ref{sec:experiments}.

\noindent\textbf{Spatial smoothness (local Huber-Laplacian).}
A local Laplacian prior over tissue fractions encourages spatial coherence across neighboring voxels:
\begin{equation}
\mathcal{R}_{\mathrm{spatial}} = \frac{\lambda_{\mathrm{sp}}}{|\mathcal{M}|} \sum_{\mathbf{x} \in \mathcal{M}} \rho\!\left(\mathbf{f}(\mathbf{x}) - \frac{1}{|\mathcal{N}(\mathbf{x})|}\sum_{\mathbf{x}' \in \mathcal{N}(\mathbf{x})} \mathbf{f}(\mathbf{x}')\right)
\end{equation}
where $\mathbf{f}$ denotes the vector of tissue fractions, $\mathcal{N}(\mathbf{x})$ is a local neighborhood (6- or 26-connected), and $\rho(\cdot)$ is the Huber loss with transition $\delta{=}0.05$.
In ablation, 26-connected smoothing is not consistently better than 6-connected smoothing. Default: $\lambda_{\mathrm{sp}}{=}0.01$.

\noindent\textbf{Topology priors.}
The core model-selection term is direction repulsion with weight $\lambda_{\mathrm{rep}}{=}0.01$:
$\mathcal{R}_{\mathrm{rep}} = \sum_{i<j} f_i f_j |{\mathbf{d}_i \cdot \mathbf{d}_j}|$.
An $L_1$ sparsity penalty on minor fiber fractions
($\tau{=}0.15$, $\lambda_{\mathrm{sparse}}{=}0.02$) is also applied.
Additional stabilization terms include orphan-WM suppression (penalizes WM fractions in voxels with near-zero $S_0$; $\lambda{=}0.01$), directional continuity (encourages neighboring voxels to share similar fiber directions; $\lambda{=}0.005$), and fiber ordering (sorts fibers by descending fraction to break permutation symmetry; $\lambda{=}0.01$).

\noindent\textbf{Calibration regularizers.}
$L_2$ penalties on per-measurement log-scales ($\alpha_n$) and offsets ($\beta_n$), and bias-field coefficients ($\mathbf{b}_\mathrm{lr}$), plus total-variation smoothness on the upsampled bias field.
These anchor all calibration parameters near identity, preventing them from absorbing biological signal.

\subsection{Optimization}
\label{sec:optimization}
PRISM optimizes within a brain mask~$\mathcal{M}$ on $b_0$-normalized signals (both $\hat{y}$ and $y$ divided by the measured $b_0$).
Two data-fidelity modes are available:

\noindent\textbf{MSE mode} treats magnitude noise as Gaussian:
\begin{equation}
\mathcal{L}_\mathrm{MSE} = \frac{1}{|\mathcal{M}|}\sum_{\mathbf{x},n}\!\big(\hat{y}(\mathbf{x},n)-y(\mathbf{x},n)\big)^2.
\label{eq:mse}
\end{equation}
This is fast and stable, requiring no noise estimate.

\noindent\textbf{Rician NLL mode} uses the correct likelihood for magnitude MRI~\cite{gudbjartsson1995rician}:
\begin{equation}
\mathcal{L}_\mathrm{NLL} = \sum_{\mathbf{x},n}\!\Big[\log\sigma^2 + \tfrac{y^2 + \hat{y}^2}{2\sigma^2} - \log I_0\!\Big(\tfrac{y\,\hat{y}}{\sigma^2}\Big)\Big],
\label{eq:nll}
\end{equation}
where $\hat{y}$ is the predicted noise-free signal from Eq.~\eqref{eq:forward}, $I_0$ is the modified Bessel function of the first kind, and $\sigma$ is the noise standard deviation.
Crucially, $\sigma$ need not be known \emph{a priori}: it is jointly optimized as part of the calibration module (Sec.~\ref{sec:artifact_model}, parameter $\sigma_\mathrm{log}$).

Both modes share the same regularization and solver.
The total objective is $\mathcal{L}_\mathrm{data} + \sum_i \lambda_i\mathcal{R}_i + \mathcal{R}_\mathrm{cal}$.

\textbf{Constraints and solver.}
Fractions via softmax; $f_\mathrm{intra}$ via sigmoid; $S_0$ via softplus; directions via $\ell_2$-normalization.
Rprop~\cite{riedmiller1993rprop} is chosen for its sign-based updates, which naturally handle heterogeneous parameter scales (orientations, fractions, calibration fields) without a global learning rate; Table~\ref{tab:optimizer} shows it reaches the lowest final MSE in the fewest iterations among eight optimizers tested on 20 axial slices.
Convergence requires 50--300 iterations depending on dataset and slab size.

\textbf{Implementation details.}
Whole-brain fitting uses overlapping 30-slice slabs (5-slice overlap) stitched via weighted averaging.
All $\lambda$ values are fixed across every experiment ($\lambda_\mathrm{sp}{=}\lambda_\mathrm{rep}{=}0.01$, $\lambda_\mathrm{sparse}{=}0.02$).
HCP whole-brain fitting uses 100 iterations per slab on a single GPU; DiSCo1 uses 300 iterations (NLL mode).
Code and configurations will be released upon acceptance.

\section{Experiments}
\label{sec:experiments}

\textbf{Datasets.}
Evaluation spans three complementary benchmarks:
(i)~\emph{Synthetic crossing-fibers}~\cite{dipy2014}: 3400 voxels (16 crossing angles from $15^\circ$ to $90^\circ$ in $5^\circ$ steps, plus single-fiber controls); 3 shells ($b{\in}\{1{,}000,\,2{,}000,\,3{,}000\}$\,s/mm$^2$, 64 directions each, $N{=}193$), SNR\,=\,30 (also 10 and 50).
A subset ($\{30,45,60,90\}^\circ$, 200 voxels each) is reused for the artifact-robustness experiment;
(ii)~\emph{DiSCo1 phantom}~\cite{rafaelpatino2021disco}: digital phantom with ground-truth connectivity ($40^3$, 16 ROIs, 120 ROI pairs), 4 shells ($b$ up to 13{,}192\,s/mm$^2$, $N{=}364$), SNR\,=\,50;
(iii)~\emph{HCP in-vivo}~\cite{vanessen2013hcp}: subject 100307, $145{\times}174{\times}145$ (${\sim}$741k brain voxels), 3 shells ($N{=}288$).

\textbf{Configuration.}
PRISM uses $K{=}2$ fibers for synthetic data, $K{=}3$ for HCP, and $K{=}5$ for DiSCo1, on a single NVIDIA RTX A6000 GPU (48\,GB VRAM).
Synthetic experiments evaluate both MSE and NLL modes with direction repulsion only (no spatial priors); NLL uses the self-supervised $\sigma$ (no oracle).  All synthetic results are seeded for deterministic reproducibility in the reported software/hardware environment.
DiSCo1 adds topology priors (300 iterations, NLL mode); HCP uses the full regularization suite (MSE mode, 100 iterations per 30-slice slab with 5-slice overlap).  For spatial smoothness, both 6-connected and 26-connected neighborhoods are evaluated in ablation.
Fiber directions are used directly as tracking peaks.

\textbf{Baselines.}
CSD~\cite{tournier2007csd}, MSMT-CSD~\cite{jeurissen2014multi}, and ODF-FP~\cite{baete2019odffp}.
CSD estimates its response function from ground-truth single-fiber voxels; MSMT-CSD receives oracle response functions matching the simulation eigenvalues.
ODF-FP uses a 1{,}000{,}000-element library with max 3 peaks, $D_a \in [1.2,2.2]$,
$D_e, D_r \in [0.1,0.6]$ ($\times 10^{-3}$\,mm$^2$/s), and DSI 8-fold tessellation.
CSD receives a single shell ($b{=}3{,}000$\,s/mm$^2$); all other methods receive the full multi-shell input.
ODF-FP is evaluated on the synthetic benchmark only; DiSCo1 comparisons use CSD and MSMT-CSD.

\textbf{Tracking \& metrics.}
Identical deterministic tracking~\cite{dipy2014} (seed density\,2, step 0.2\,mm, fixed seed).
Connectivity scored via Pearson~$r$; angular error reported as best-match (each ground-truth fiber matched to its closest predicted fiber).
F1 and recall quantify fiber detection.

\section{Results}

\paragraph{Robustness to Intensity Nuisance Factors.}
\label{sec:artifact_results}
Nuisance calibration is evaluated under per-measurement gain perturbations ($\sigma_g{=}0.02$--$0.20$, SNR\,=\,30, crossing angles $\{30,45,60,90\}^\circ$, 200 voxels each).
At $\sigma_g{=}0.20$, calibration halves angular error ($4.8^\circ{\to}2.4^\circ$, $-$50\%) and removes 85\% of reconstruction MSE ($7.6{\to}1.1{\times}10^{-3}$);
on clean or well-preprocessed data, all calibration parameters self-regularize to identity (see ablation below).

\paragraph{Crossing-Fiber Benchmark.}
Table~\ref{tab:multitensor_sim_snr30} reports fiber recovery on the 16-angle synthetic benchmark (Sec.~\ref{sec:experiments}).
PRISM$_{\text{MSE}}$ achieves $3.5^\circ$ overall best-match angular error with 95\% recall, $1.9{\times}$ lower than the best baseline (MSMT-CSD, $6.8^\circ$, 83\% recall).
In NLL mode (learned~$\sigma$), error drops to $2.3^\circ$ with 99\% recall and 99\% F1.
At narrow crossings ($\leq 30^\circ$), CSD and MSMT-CSD largely detect only a single fiber (overall recall 82--83\%), while ODF-FP partially resolves the second peak and achieves the lowest baseline error in this regime ($7.4^\circ$--$13.7^\circ$ vs.\ $7.8^\circ$--$15.2^\circ$ for MSMT-CSD); however, ODF-FP's dictionary discretization degrades at wider angles (${\geq}45^\circ$, Table~\ref{tab:multitensor_sim_snr30}).
PRISM$_{\text{NLL}}$ resolves both fibers across all angles with $3.1^\circ$--$5.8^\circ$ best-match error at $\leq 30^\circ$.
PRISM uses a deliberately mismatched forward model ($\lambda_\perp{=}0.4$ vs.\ true $0.3{\times}10^{-3}$\,mm$^2$/s), while baselines receive oracle or ground-truth-derived response functions (Sec.~\ref{sec:experiments}).


\begin{table}[t]
\centering
\caption{Best-match angular error (\textdegree) at SNR\,=\,30 (16 angles, 15--90\textdegree; representative subset). F1 and recall (\%) over all angles.}
\label{tab:multitensor_sim_snr30}
\footnotesize
\setlength{\tabcolsep}{2.5pt}
\setlength{\aboverulesep}{2pt}
\setlength{\belowrulesep}{2pt}
\resizebox{\textwidth}{!}{%
\begin{tabular}{@{}lcccccccccccc@{}}
\toprule
Method & 1-fib & $15^\circ$ & $20^\circ$ & $25^\circ$ & $30^\circ$ & $45^\circ$ & $60^\circ$ & $75^\circ$ & $90^\circ$ & Overall & F1 & Rec. \\
\midrule
PRISM$_{\mathrm{MSE}}$ & \textbf{0.7} & 7.5 & 9.9 & 11.7 & 8.2 & 1.9 & 1.5 & 1.5 & \textbf{1.3} & 3.5 & 96 & 95 \\
PRISM$_{\mathrm{NLL}}$ & 1.4 & \textbf{5.8} & \textbf{4.9} & \textbf{3.9} & \textbf{3.1} & \textbf{1.9} & \textbf{1.5} & \textbf{1.3} & \textbf{1.3} & \textbf{2.3} & \textbf{99} & \textbf{99} \\
\midrule
CSD & 3.1 & 7.9 & 10.3 & 12.7 & 15.2 & 8.8 & 3.9 & 4.0 & 3.5 & 7.7 & 90 & 82 \\
MSMT-CSD & 3.0 & 7.8 & 10.2 & 12.7 & 15.2 & 5.3 & 3.4 & 3.2 & 3.1 & 6.8 & 91 & 83 \\
ODF-FP & 3.1 & 7.4 & 9.7 & 11.9 & 13.7 & 16.9 & 10.9 & 9.4 & 10.9 & 11.6 & 87 & 86 \\
\bottomrule
\end{tabular}}
\end{table}



\begin{table}[t]
\centering
\caption{Optimizer comparison (200 iterations, Stanford HARDI, 20 middle axial slices,
  $n{=}3$ fibers, seed-42 init).
  Convergence threshold (Iter$\to$thr): iteration first reaching 110\% of best final MSE.
  Rprop achieves the lowest final MSE and converges fastest.}
\label{tab:optimizer}
\footnotesize
\setlength{\tabcolsep}{2pt}
\setlength{\aboverulesep}{2pt}
\setlength{\belowrulesep}{2pt}
\resizebox{\textwidth}{!}{%
\begin{tabular}{@{}lcccccccc@{}}
\toprule
 & Rprop & Adam & Adam\,(.05) & AdamW & NAdam & RMSprop & Adagrad & SGD+m \\
\midrule
MSE ($\!\times\!10^{-3}$) & \textbf{21.2} & 24.4 & 21.5 & 24.4 & 24.5 & 21.5 & 45.7 & 85.3 \\
Iter$\to$thr & \textbf{22} & $>$200 & 63 & $>$200 & $>$200 & 39 & $>$200 & $>$200 \\
\bottomrule
\end{tabular}}
\end{table}



\paragraph{DiSCo1 Tractography.}
On the DiSCo1 phantom (16 ROIs, 120 ROI pairs, 11{,}528 ground-truth strands, SNR\,=\,50), we run identical deterministic tracking for CSD, MSMT-CSD, and PRISM (with Rician NLL) and evaluate connectivity correlation at tracking angles 15--30$^\circ$.
PRISM outperforms both baselines at \emph{every} evaluated angle:
at their optimal angles, CSD reaches $r{=}.917$ (30$^\circ$), MSMT-CSD reaches $r{=}.920$ (30$^\circ$), and PRISM achieves $r{=}.934$ (25$^\circ$).
Across the full range, PRISM scores $r{=}.925/.930/.934/.928$ versus $.906/.911/.918/.920$ for MSMT-CSD, with the largest per-angle gains at 15$^\circ$ and 20$^\circ$ ($+1.9$\,pp).
PRISM's optimum shifts to a tighter tracking angle (25$^\circ$ vs.\ 30$^\circ$), indicating that its fiber orientations support narrower angular thresholds while maintaining higher overall connectivity.

\paragraph{In-Vivo Whole-Brain Fitting.}
On HCP data (741k voxels, 288 measurements, 3 shells), PRISM reaches MSE\,=\,$6.3{\times}10^{-3}$ in $\sim$12\,min (MSE mode; hardware in Sec.~\ref{sec:experiments}).
The recovered tissue fractions are spatially consistent with known anatomy: cortical GM ribbons, deep WM tracts, and ventricular CSF.


Compared against FreeSurfer $T_1$ segmentations, PRISM achieves Dice\,${=}\,0.79$ (WM) and $0.80$ (GM) with Pearson~$r{=}0.75$/$0.67$; all summary metrics vary by ${<}0.003$ across three random seeds.

\paragraph{Ablation.}
\label{sec:ablation}
We ablate PRISM on DiSCo1 ($r{=}.934$ full model).
The restricted compartment contributes the largest single gain: removing it drops $r$ to $.840$ ($-9.4$\,pp).
Removing the topology prior yields $r{=}.915$ ($-1.9$\,pp); removing the Huber--Laplacian spatial prior yields $r{=}.927$ ($-0.7$\,pp).
Nuisance calibration has no measurable effect on this unperturbed phantom ($\Delta r < 0.1$\,pp) or on well-preprocessed HCP data, consistent with self-regularization to identity.


\section{Discussion and Conclusion}

PRISM reframes microstructure estimation as a differentiable fitting problem, coupling an explicit multi-compartment forward model with end-to-end optimization rather than isolated voxelwise fitting stages; this explicit fiber parameterization yields a $1.9{\times}$ angular-error reduction over the strongest baseline on the 16-angle benchmark (Table~\ref{tab:multitensor_sim_snr30}), because each fiber is treated as a first-class parameter rather than an implicit SH peak.
The ablation shows the restricted compartment provides the largest single gain ($-9.4$\,pp when removed), with topology ($-1.9$\,pp) and spatial priors ($-0.7$\,pp) contributing further.
The intensity perturbation study confirms nuisance calibration activates under gain corruption while self-regularizing to identity on unperturbed data.
On HCP, dMRI-derived tissue fractions agree with FreeSurfer (Dice$\approx 0.79$/$0.80$ WM/GM) without T1 input.

\textbf{Limitations.}
PRISM currently models intensity-domain nuisance factors; geometric preprocessing (motion, susceptibility) is still assumed. 
Convergence could be accelerated by replacing random initialization with a self-supervised encoder or by warm-starting from a dictionary-based method (e.g., ODF-FP~\cite{baete2019odffp}, FORCE~\cite{shah2025force}), reducing sensitivity to local minima.
Further extensions include Bingham dispersion and geometric distortion operators.

In summary, PRISM demonstrates that differentiable analysis-by-synthesis with explicit fibers, restricted diffusion, and learned-$\sigma$ Rician fitting yields substantial gains in crossing resolution, tractography agreement, and anatomically consistent whole-brain maps on a single GPU.


\begingroup
\emergencystretch=3em
\hbadness=10000          
\bibliographystyle{splncs04}
\bibliography{references}
\endgroup

\end{document}